\documentclass[lettersize,journal]{IEEEtran}
\usepackage{amsmath,amsfonts}
\usepackage{algorithmicx}
\usepackage{algorithm}
\usepackage{array}
\usepackage[caption=false,font=normalsize,labelfont=sf,textfont=sf]{subfig}
\usepackage{textcomp}
\usepackage{stfloats}
\usepackage{CJKutf8}
\usepackage{url}
\usepackage{color}
\usepackage{algpseudocode}
\usepackage{xcolor}
\usepackage{makecell}
\usepackage[hidelinks]{hyperref}
\usepackage{verbatim}
\usepackage{graphicx}
\usepackage{cite}
\usepackage{booktabs}
\usepackage{multirow}
\begin{document}
\begin{CJK}{UTF8}{gbsn}

\title{OverlapMamba: Novel Shift State Space Model for LiDAR-based Place Recognition}

\author{	Qiuchi Xiang,
Jintao Cheng,
Jiehao Luo,
Jin Wu,~\IEEEmembership{Member,~IEEE,}
Rui Fan,~\IEEEmembership{Senior Member,~IEEE,}
Xieyuanli Chen,~\IEEEmembership{Member,~IEEE,}
Xiaoyu Tang,~\IEEEmembership{Member,~IEEE,}
\thanks{This research was supported by the National Natural Science Foundation of China under Grants 62001173 and 62233013, the Project of Special Funds for the Cultivation of Guangdong College Students’ Scientific and Technological Innovation (``Climbing Program'' Special Funds) under Grants pdjh2022a0131 and pdjh2023b0141, the Science and Technology Commission of Shanghai Municipal under Grant 22511104500, the Fundamental Research Funds for the Central Universities, and the Xiaomi Young Talents Program. \textit{Qiuchi Xiang and Jintao Cheng are cofirst authors. })(\textit{Corresponding author: Xiaoyu Tang}).}

\thanks{Qiuchi Xiang and Jiehao Luo are with the School of Data Science and Engineering, Xingzhi College, South China Normal University, Shanwei, 516600, China (e-mail: 20218131007@m.scnu.edu.cn; 20228132034@m.scnu.edu.cn) }
\thanks{Jintao Cheng and Xiaoyu Tang are with the School of Electronic and Information Engineering, South China Normal University, Foshan, Guangdong 528225, China (e-mail: chengjt\_alex@outlook.com; tangxy@scnu.edu.cn)}
\thanks{Jin Wu is with the Department of Electronic and Computer Engineering, Hong Kong University of Science and Technology, Hong Kong, China (e-mail: jin\_wu\_uestc@hotmail.com) }
\thanks{Xieyuanli Chen is with the College of Intelligence Science and Technology, National University of Defense Technology, Changsha, China. {\tt\small xieyuanli.chen@nudt.edu.cn}}
\thanks{Rui Fan is with the College of Electronics \& Information Engineering, Shanghai Research Institute for Intelligent Autonomous Systems, the State Key Laboratory of Intelligent Autonomous Systems, and Frontiers Science Center for Intelligent Autonomous Systems, Tongji University, Shanghai 201804, China. {\tt\small rui.fan@ieee.org}}

\thanks{Manuscript received April 19, 2021; revised August 16, 2021. }}

\markboth{Journal of \LaTeX\ Class Files,~Vol.~14, No.~8, August~2021}%
{Shell \MakeLowercase{\textit{et al.}}: A Sample Article Using IEEEtran.cls for IEEE Journals}


\maketitle

\begin{abstract}
Place recognition is the foundation for enabling autonomous systems to achieve independent decision-making and safe operations. It is also crucial in tasks such as loop closure detection and global localization within SLAM. Previous methods utilize mundane point cloud representations as input and deep learning-based LiDAR-based Place Recognition (LPR) approaches employing different point cloud image inputs with convolutional neural networks (CNNs) or transformer architectures. However, the recently proposed Mamba deep learning model, combined with state space models (SSMs), holds great potential for long sequence modeling. Therefore, we developed OverlapMamba, a novel network for place recognition, which represents input range views (RVs) as sequences. In a novel way, we employ a stochastic reconstruction approach to build shift state space models, compressing the visual representation. Evaluated on three different public datasets, our method effectively detects loop closures, showing robustness even when traversing previously visited locations from different directions. Relying on raw range view inputs, it outperforms typical LiDAR and multi-view combination methods in time complexity and speed, indicating strong place recognition capabilities and real-time efficiency.
\end{abstract}

\section{Introduction}

\begin{figure}[!t]
    \centering
    \includegraphics[width=0.9\linewidth]{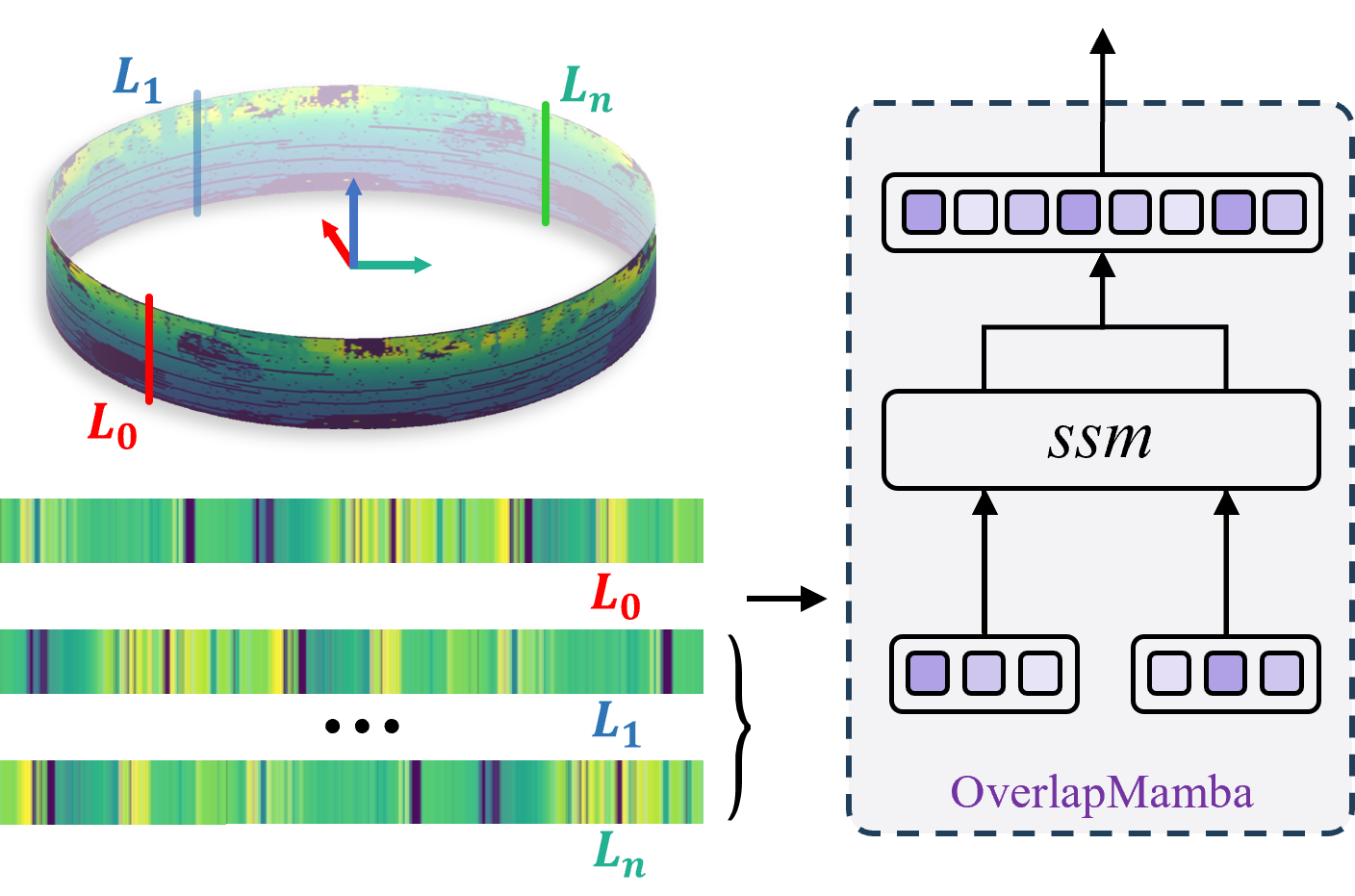}
\caption{Core idea of the proposed OverlapMamba model. The left parts represent RV projection and 1D point cloud serialization. The right parts represent the overview of our novel state space models for place recognition.}
\label{fig}
\end{figure}

\IEEEPARstart{P}{lace} recognition \cite{r1, r2} is a key technology in the process of simultaneous localization and mapping (SLAM) of autonomous driving or unmanned systems. This technology matches the scene captured by the sensor with the scene in the database through feature comparison, determining the global localization of the autonomous driving system. Place recognition technology is one of the key technologies that guarantees the ability of the autonomous system to navigate unfamiliar environments \cite{Rasouli_Kotseruba_Tsotsos_2018}. Therefore, researching and developing more accurate and robust place recognition algorithms has become an important research direction in SLAM \cite{Bresson_Alsayed_Yu_Glaser_2017}. In this paper, we propose a novel place recognition method that utilizes 3D LiDARs installed on autonomous driving systems to generate original range views (RVs). Subsequent operations are applied to these RVs, yielding a robust global descriptor for recognizing locations from different viewpoints in outdoor driving scenarios.

Place recognition technology enables autonomous systems to perceive their surroundings without prior knowledge, accurately identifying their position in complex and dynamic environments. Contemporary image-based positioning techniques usually involve regressing image similarities or matching descriptors to retrieve the most similar reference scans from databases. However, these methods can result in inaccurate positioning when autonomous systems revisit previously mapped locations, as dynamic environmental changes and lighting variations can affect accuracy. In contrast, LPR methods exhibit greater robustness and competitive advantages due to their lower susceptibility to environmental perturbations. These methods utilize 3D LiDAR for data acquisition and employ diverse data representations, including Bird's Eye Views \cite{r4}, RVs \cite{r20}, and raw 3D point clouds \cite{r24}, as inputs for LPR. For instance, RangeRCNN \cite{r51} introduces an RV-PV-BEV (range-point-bird's eye views) module that transfers features from RVs to BEVs, circumventing issues related to scale variation and occlusion. Nonetheless, the computational intensity of aggregating multiple images for global descriptor generation renders these methods unsuitable for rapid computation and real-time operations. Consequently, graph-based online operational methods incur significant delays, and as maps expand, the time required for location identification and loop closure detection proportionally increases.

This study introduces an end-to-end network that utilizes RVs as input to generate yaw-invariant global descriptors for each location to address the high latency inherent in existing graph-based online operation methods. This robust place recognition approach is achieved by matching global descriptors across multiple viewpoints. Furthermore, this work presents OverlapMamba block, a novel feature extraction backbone, which focuses on integrating the cutting-edge state space model (SSM) into SLAM techniques to enhance the efficiency of place recognition and bolster global localization abilities. The OverlapMamba block models RVs in a multi-directional sequence and computes attention by employing a self-designed random reconstruction method that compensates for the unidirectional modeling and positional awareness limitations of SSM. A straightforward sequential pyramid pooling (SPP) architecture is integrated within the network to mitigate feature loss from noise interference. 

Compared with the most convincing transformer-based LPR method \cite{r30}, OverlapMamba outperforms the transformer architecture by delivering superior visual representations with linear computational complexity. The resultant global descriptor demonstrates robustness in place recognition tasks due to its yaw-invariant architecture and the incorporation of spatially rich features, maintaining efficacy even when the autonomous system navigates in reverse.

The method in this paper evaluates the closed-loop detection and place recognition performance of the proposed lightweight network on three public datasets. The experimental results show that our proposed method is superior to the most advanced methods in both closed-loop detection and place recognition, which proves that our method is highly competitive. In summary, we make three claims regarding our approach:
\begin{enumerate}
\item{A backbone network for RV multidirection sequence modeling is proposed that can generate multiscale yaw-invariant feature maps for the subsequent generation of global descriptors to extract location information in place recognition tasks.}
\item{A random reconstruction method is innovatively designed to process RVs. Additionally, the OverlapMamba block is used to enhance the attention of the multi-scale feature map extracted from the backbone. We also design a simple SPP architecture to denoise  the feature graph and  enhance the robustness of the network. Relying on raw range view inputs, our method outperforms typical LiDAR and multi-view combination methods in time complexity and speed.}
\item{The proposed method achieves SoTA on the loop closure detection task of the KITTI dataset, and the performance and generalizability of the proposed method on the place recognition task are validated in benchmark tests of two other public datasets. Our OverlapMamba has been implemented at \href{http://github.com/SCNU-RISLAB/OverlapMamba}{http://github.com/SCNU-RISLAB/OverlapMamba}}.
\end{enumerate}

\section{Related work}
In the early stages of autonomous driving, scientists conducted extensive work on visual place recognition (VPR) using cameras as the primary sensors \cite{r31}. Thus, the understanding of VPR research is primarily based on the review literature by Lowry et al. \cite{r39}. Here, we focus on the related work of LPR.

\subsection{LPR Based on Local Description}
Previous methods for LPR primarily generated local descriptions through manual crafting or deep learning approaches. These methods effectively capture unique features such as texture and color, representing the context of their surroundings. Initially, inspired by spin image \cite{r40}, manually crafted local descriptions were commonly used for LPR tasks, such as recognizing and verifying data from the same site collected at different times. This was achieved by extracting keypoints from the geometric and topological structures of point clouds and calculating descriptors for these points manually to match between point clouds. For instance, Jiaqi Yang et al. \cite{r42} introduced a method for generating local descriptions through weighted projection vectors to enhance the stability of LPRs. Fengkui Cao et al. \cite{r43} proposed an image model named the bearing angle (BA), which extracts features for scene matching. Another approach generated local descriptions through deep learning, typically encoding point cloud patches with 3D CNNs. Charles R. Qi et al. \cite{r24} designed a method that directly utilized point clouds, using their permutation invariance to produce efficient and robust local descriptions. However, both local description-based LPR methods are susceptible to viewpoint changes affecting keypoint accuracy and relying on substantial computational power to process dense point clouds; thus, these methods face limitations in handling sparse point clouds from high-precision LiDAR devices.

\subsection{LPR Based on Global Description}
Recent methods have favored the use of popular global description-based approaches to describe the overall scene features, providing a holistic view of the data. These methods typically use various forms of data as input, such as RV, BEV, and spherical view. Xieyuanli Chen et al. \cite{r29} proposed a network that can address loop closure detection and scene recognition. This approach intuitively and efficiently estimates the similarity between scan pairs by overlapping distance images. Subsequently, OverlapTransformer \cite{r30} was introduced as an enhanced version of the previous model. This lightweight network leverages the transformer architecture to incorporate attention weighting on yaw-invariant descriptors, resulting in a notable improvement in location recognition performance. Building on OverlapTransformer, Junyi Ma et al. \cite{r45} proposed a cross-view transformer network that fused RVs and BEVs generated from LiDAR data, improving the robustness of global descriptors. Other methods \cite{r20} also adopted the transformer architecture, which is known for capturing long-range dependencies and contextual relationships, to achieve effective recognition in cluttered environments. However, their substantial computational demands limit batch sizes during training. Methods \cite{r51} employ projection-based approaches, which offer lower computational needs and better interpretability but inevitably lose information during dimensionality reduction.

\begin{figure*}[htbp]
    \centering
\vspace{+.3cm}
    \includegraphics[width=0.9\linewidth]{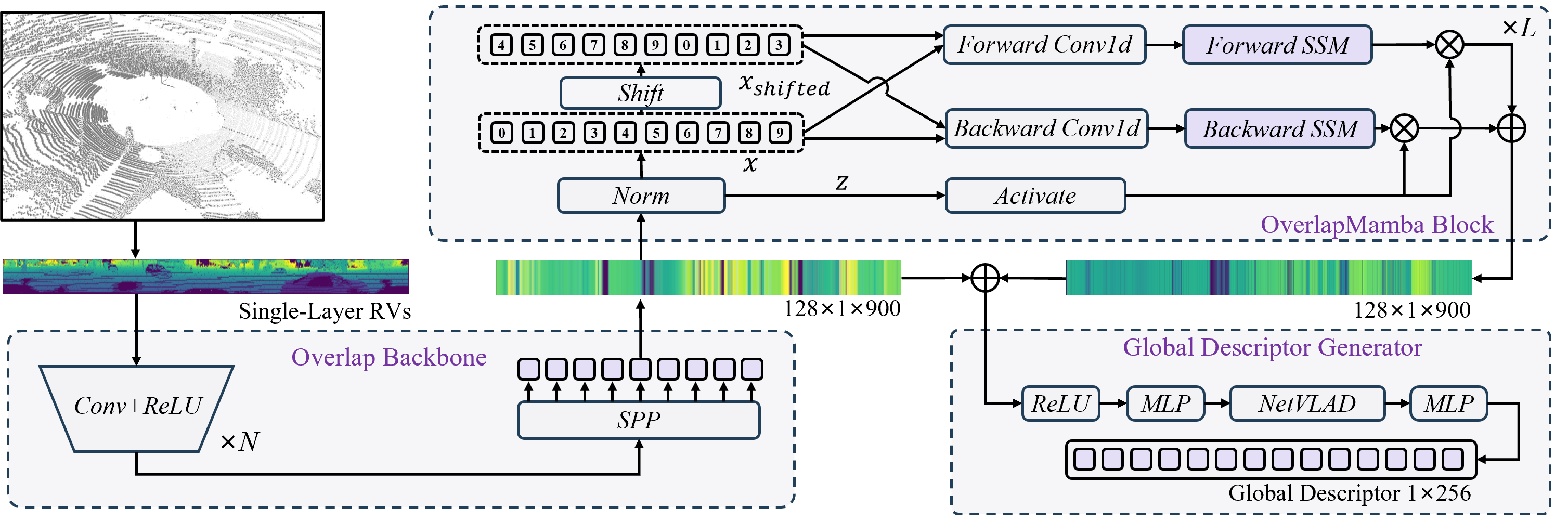}
\caption{Overview of the proposed OverlapMamba. Assuming a batch size of 1, the overlap backbone compresses the RVs from the LiDAR sensor information into yaw-equivariant feature sequences. The OverlapMamba block connects the feature sequences from the backbone with the multidirectionally enhanced feature sequences processed by the SSM. The global descriptor generator (GDG) utilizes a combination of multilayer peceptron (MLP) and NetVLAD to generate a one-dimensional global descriptor.}
\label{fig1}
\vspace{-.3cm}
\end{figure*}

\section{Overview of the Framework}\label{sec3}

This study is focused on integrating the cutting-edge SSM, known as the Mamba model, into SLAM techniques to enhance the efficiency of place recognition and bolster the capabilities of global localization. We first describe the preliminaries of SSMs. In Section~\ref{sec3.2}, we introduce the concept of OverlapMamba, with its comprehensive framework depicted in Figure~\ref{fig1}. Then, in Section~\ref{sec3.3}, we detail the architecture of the OverlapMamba block and illustrate how to model the input sequence. Given that the global descriptors serving as inputs to Mamba are convolved from RVs along the vertical dimension, we elaborate on sequence pyramid pooling for sequences in Section~\ref{sec3.4}, which performs multiscale one-dimensional pooling on sequences to preserve spatial information. In Section~\ref{sec3.5}, we discuss the issue of using triplet loss with computed overlap labels during training.

\subsection{Preliminaries}\label{sec3.1}

The structured state space model (S4) based on SSM and Mamba \cite{mamba} is inspired by continuous systems, which map a 1-D function or sequence \(x(t) \in \mathbb{R}\) to \(y(t) \in \mathbb{R}\) through a hidden state \(h(t) \in \mathbb{R}^N\). Mathematically, they are often formulated as linear ordinary differential equations (ODEs), with parameters including \(A \in \mathbb{R}^{N \times N}\), \(B, C \in \mathbb{R}^N\), and skip connection parameter \(D \in \mathbb{R}^1\). In this system, \(A\) serves as the evolution parameter, while \(B\) and \(C\) act as projection parameters.
\begin{equation}
\begin{aligned}
h'(t) = Ah(t) + Bx(t)\\
y(t) = Ch(t) + Dx(t)
\end{aligned}
\end{equation}

As continuous-time models, SSMs face significant challenges when integrated into deep learning algorithms. Discretization is necessary to overcome this obstacle, and S4 and Mamba are discrete versions of continuous systems with the following discretization rules.

\begin{equation}
\begin{gathered}
\bar{A} = e^{\Delta A}, \\
\bar{B} = (e^{\Delta A} - I) A^{-1} B, \\
h_k= \bar{A} h_{k-1} + \bar{B} x_k, \\
y_k = C h_k + D x_k, \\
\end{gathered}
\end{equation}

The following formula is used to enable parallel training and derive the kernel for efficient computation of \(y\) using convolution: \(M\) represents the sequence length, and \(\bar{K}\) represents the kernel of the 1D convolution.

\begin{equation}
\begin{aligned}
C h_k& =C\left(\bar{A} h_{k-1}+\bar{B} x_{k}\right) \\
& =C\left(\bar{A}\left(\bar{A} h_{k-2}+\bar{B} x_{k-1}\right)+\bar{B} x_k\right) \\
& =C  \bar{A}^k  \bar{B} x_0+C  \bar{A}^{k-1}  \bar{B}  x_1+...+C  \bar{B} x_k
\end{aligned}
\end{equation}

\begin{equation}
\begin{gathered}
\bar{K} = (C\bar{B}, C\overline{AB}, \ldots, CA^{M-1}\bar{B})\\
y = x\bar{K}+ D x
\end{gathered}
\end{equation}

\subsection{Mamba-Based Place Recognition}\label{sec3.2}

An overall overview of the OverlapMamba model, consisting of the overlap backbone, OverlapMamba block, and the final GDG, is shown in Figure~\ref{fig1}. We use point cloud data generated by raw LiDAR scans and create RVs from them. Projection transformation between point clouds and RVs is necessary. The point cloud \(P\) can be projected onto the RVs \(R\) through \(\Pi: \mathbb{R}^3 \rightarrow \mathbb{R}^2\), where each 3D point is transformed into a pixel on \(R\). Each point \(p_k = (x, y, z)\) is transformed into image coordinates \((u, v)\) as follows:

\begin{equation}
\left(\begin{array}{l}
u_k \\
v_k
\end{array}\right)=\left(\begin{array}{c}
\frac{1}{2}\left[1-\arctan (y_k, x_k)/ \pi\right] w \\
{\left[1-\left(\arcsin \left(z_k/r_k\right)+{f}_{{up}}\right)/ {f}\right] h}
\end{array}\right)
\end{equation}

where \(r_k = ||p_k||_2 \) is the distance measurement for the corresponding point \(p_k\), \(f = f_{\text{up}} + f_{\text{down}}\) is the vertical field of view of the sensor, and \(w, h\) are the width and height of the resulting RVs, respectively.

We use single-channel RVs (assuming a batch size of 1) with a size of \(1 \times h \times w\). Single-channel RVs, which provide more straightforward depth information, offer a more memory-efficient alternative during training compared to three-channel RGB images. Originally designed for 1D sequences, the standard Mamba model is adapted for visual tasks by converting RVs into a sequential data format. We adopt the idea from OverlapLeg~\cite{r29}, where in the backbone, we use only convolutional filters along the vertical dimension without compressing the width dimension. In OverlapLeg, the RV is split into sequences with a size of \(h\times1\) along the vertical dimension, processed with 1-D convolution, and then concatenated into a single \(1\times w\) sequence. However, the single-channel RV inevitably does not have enough spatial information to ensure that the final generated sequence does not lose scene information. The error is caused by noise amplification due to single-dimensional processing. Therefore, we introduce sequence pyramid pooling in our backbone design. 

Inspired by vision mamba~\cite{zhu2024vision}, we use Mamba to process sequences with high accuracy and efficiency. The standard Mamba is tailored for one-dimensional sequences. We serialize the RVs to handle visual tasks into \(x \in \mathbb{R}^{c \times 1 \times w}\), where \(c\) is the number of channels and \(w\) is the width of the RVs. Then, we send \(x_{l-1}\) to the \(l\)-th layer of the OverlapMamba encoder to obtain the output \(x_l\). Finally, we apply an activation function to the output \(x_l\), normalize it, and pass it backward to the GDG.

In the GDG, we use NetVLAD~\cite{r33} to generate yaw-invariant descriptive feature symbols. NetVLAD facilitates end-to-end image-based place recognition with inherent invariance to yaw rotation. For example, if the input raw LiDAR data are rotated 90 degrees and 180 degrees, the distance in the distance image will shift by \(\frac{1}{4}w\) and \(\frac{1}{2}w\), respectively. However, the final generated global descriptor is yaw-invariant so that it will generate the same global descriptor in both cases. The overall process is shown in the following formula.
\begin{equation}
\begin{gathered}
	{x}_l={Olm}\left( {x}_{l-1} \right) +{x}_{l-1},\\
	{n}_{l}={Norm}\left( {x}_l \right) ,\\
	\hat{g}={G}({n}_{l})\\
\end{gathered}
\end{equation}


where \(Olm(.)\) denotes the OverlapMamba block, and the input sequence \(x_{l-1}\) is concatenated with itself after passing through the OverlapMamba block. \(G(.)\) represents the GDG, which is responsible for converting the standardized sequence into the final global descriptor.

\subsection{OverlapMamba block}\label{sec3.3}

\begin{figure}[!t]
    \centering
    \vspace{+.3cm}
    \includegraphics[width=0.9\linewidth]{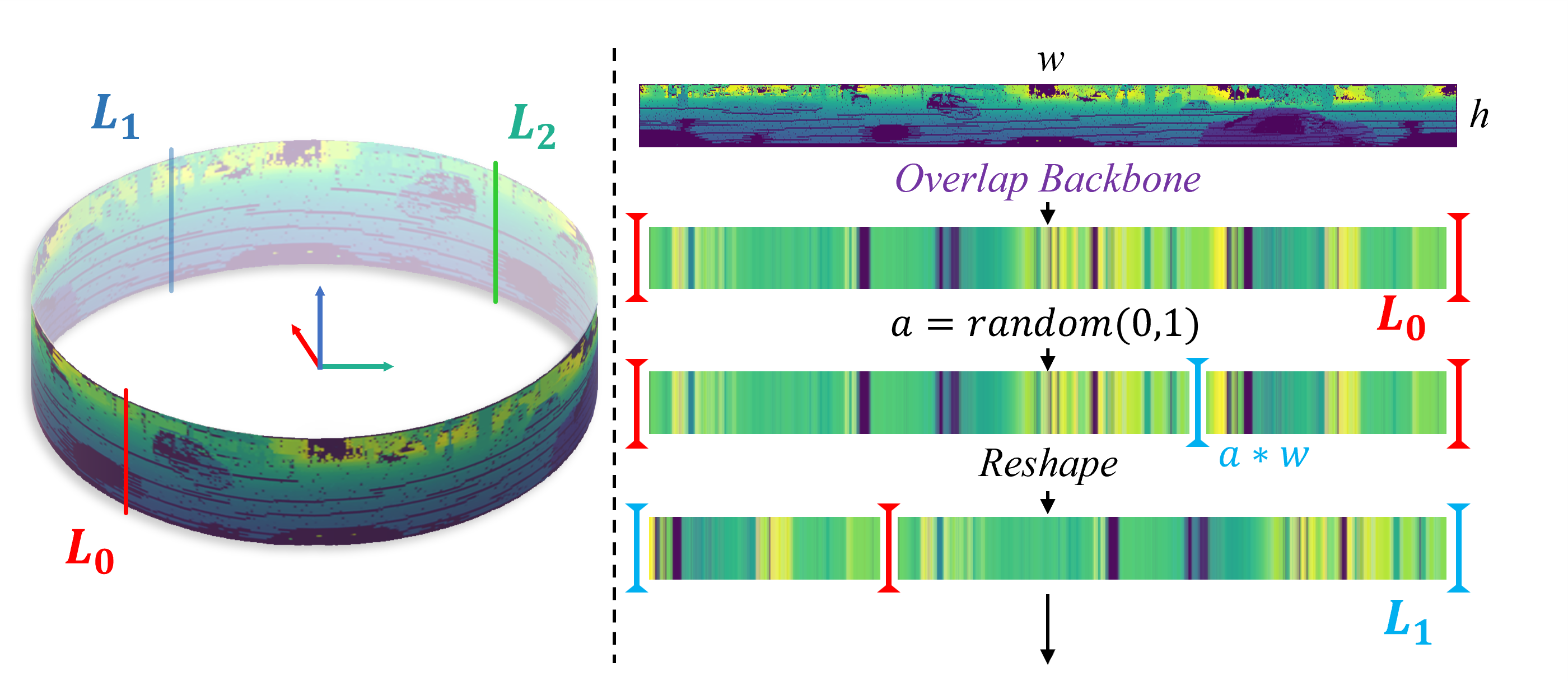}
\caption{The SIFT operation process. The left part shows an example of an RV containing omnidirectional feature information. In the right part of the figure, we demonstrate the process of randomly reconstructing the feature sequence modeled along the vertical direction for the yaw angle, where $a$ is a random parameter used to calculate the starting index of the reconstructed sequence.}
\label{fig2}
\vspace{-.3cm}
\end{figure}

The proposed OverlapMamba block (OLM) is illustrated in Figure~\ref{fig1}. The original Mamba module is specifically designed for 1D sequences and may not be suitable for tasks requiring spatial awareness. In recent research, a prevalent approach among researchers has been to employ bidirectional sequence modeling~\cite{zhu2024vision}. This method essentially divides the image into multiple patches and incorporates positional encoding to map them into sequences. Concurrently, some studies opt for using four distinct directional sequences as input~\cite{ruan2024vm, liu2024vmamba}, gathering pixel information along both vertical and horizontal axes and then reversing these generated sequences to create a quartet of sequences. Finally, after passing through the selective SSM (S6) model \cite{mamba}, all sequences are merged into a new sequence.

\begin{algorithm}[t]
\definecolor{mygreen}{RGB}{0,100,0}
\caption{\enskip \textbf{OverlapMamba block Process}}\label{alg1}
\begin{algorithmic}
    \State \textbf{Input}: token sequence $\text{T}_{l-1}$: \textcolor{mygreen}{(B,M,D)} 
    \State \textbf{Output}: token sequence $\text{T}_{l}$: \textcolor{mygreen}{(B,M,D)} 
    \State/* normalize the input sequence $\text{T}_{l-1}$ */
    \State $\text{T}_{l-1}^{'}$: \textcolor{mygreen}{(B,M,D)}  $\gets$ Norm($\text{T}_{l-1}$)
    \State x, z: \textcolor{mygreen}{(B,M,E)}  $\gets$ Linear$^{\text{x}}(\text{T}_{l-1}^{'})$, Linear$^{\text{z}}(\text{T}_{l-1}^{'})$
    \State
    \State/* process with random yaw and different directions */
    \State x$_{\text{shift}}$:\textcolor{mygreen}{(B,M,E)}$\gets$ Shift(x)
    \State x$_{\text{backward}}$, x$_{\text{shift\_backward}}$: \textcolor{mygreen}{(B,M,E)}$\gets$ Flip(x), Flip(x$_{\text{shift}}$)
    \For {$o$ in \{different directions\}}
    \State x$_{o}^{'}$: \textcolor{mygreen}{(B,M,E)} $\gets$ SiLU(Conv1d$_{o}$(x))
    \State A$_o$ $\gets$ exp(Parameter$_{o}^{A}$)
    \State D$_o$ $\gets$ Parameter$^{D}_{o}$
    \State $\varDelta ,\ \text{B}_o,\ \text{C}_o\ \gets \ \text{Split}\left( \text{Linear}\left(       \text{x}_o^{\text{$^{'}$}} \right) \right)$
    \State  $\varDelta_{o}$:\textcolor{mygreen}{(B,M,E)} $\gets $ Softplus(Linear$_{o}^{\varDelta}$($\varDelta$))
    \State$\overline{\text{A}}_o,\ \overline{\text{B}}_o: \text{\textcolor{mygreen}{(B,M,E,N)}}  \gets \Delta _o\otimes \text{A}_o,\ \Delta _o\otimes \text{B}_o$
    \State$\text{y}_o:$ \textcolor{mygreen}{(B,M,E)} $ \gets \text{SSM}\left( \overline{\text{A}}_o,\overline{\text{B}}_o,\text{C}_o,\text{D}_o \right) \left( \text{x$^{'}$}_o \right) $    
    \EndFor
    \State
    \State/* get y$^{'}$ */
    \For {$i$ in \{different directions\}}
        \State$\mathrm{y}_{i}^{'}:\text{\textcolor{mygreen}{(B,M,D)} }\gets \mathrm{y}_i\odot \text{SiLU}\left( \mathrm{z} \right)$
    \EndFor   
    \State
    \State/* residual connection */
    \State$\text{T}_l: \text{\textcolor{mygreen}{(B,M,D)} }  \gets \text{Linear}^T\left( \text{Sum}\left( \text{y$^{'}$} \right) \right) +\text{T}_{l-1}$
    \State Return: T$_l$
\end{algorithmic}
\end{algorithm}

In this paper, the convolutional filters in the overlap backbone compress only the range image along the vertical dimension without compressing the width dimension. This results in a feature sequence with a maximum output size of \(1 \times w \times c\). We adopt a bidirectional approach for sequence modeling. Adding additional positional embeddings or sampling along the horizontal direction is unnecessary since we directly obtain token sequences through the stack of convolutional modules. The token sequences directly contain yaw information, and flipping the reverse sequence after processing contains information about the robot approaching from the opposite direction in the same scene. Therefore, we believe that due to the global scene information contained in the distance image, token sequences are generated at different yaw angles in the same scene from a cyclic sequence, as shown in Figure~\ref{fig2}. Therefore, in the overlap backbone, we use the \(Shift(\cdot )\) function from Algorithm~\ref{alg1} to randomly process the normalized token sequences with yaw angles and generate sequences with randomly flipped yaw angles. The processed data can simulate the features of the same scene under different yaw angles, enhancing the generalization ability of the model during training. Finally, after processing, four different sequences are obtained as inputs to the selective SSM (S6) for inference and training.

The overall OverlapMamba block combines multidirectional sequence modeling for place recognition tasks. We demonstrate the operations of the OLM block in Algorithm 1, with the following hyperparameters: \(L\) for the number of module stackings, \(D\) for the hidden state dimension, \(E\) for the extended state dimension, and \(N\) for the SSM dimension. The block receives and normalizes the token sequence \(T_{l-1}\); then, it uses a linear layer to project the sequence to obtain \(x\) and \(z\). Next, we flip \(x\) and apply random yaw angle processing to obtain four directional sequences, each of which is processed separately. We pass through a 1D convolution and activation function for each sequence to obtain \(x'_o\). Then, we slice the result of the linear layer. Using the softplus($\cdot$) function separately for \(\varDelta\), we calculate \(\overline{A}\) and \(\overline{B}\) and input them into the SSM to obtain \(y'\) gated by \(z\). Finally, the output token sequence \(T_l\) is obtained by adding the sequences from the four directions.

\subsection{Sequential Pyramid Pooling in the Backbone}\label{sec3.4}

\begin{figure}[!t]
    \centering
    \vspace{+.3cm}
    \includegraphics[width=0.8\linewidth]{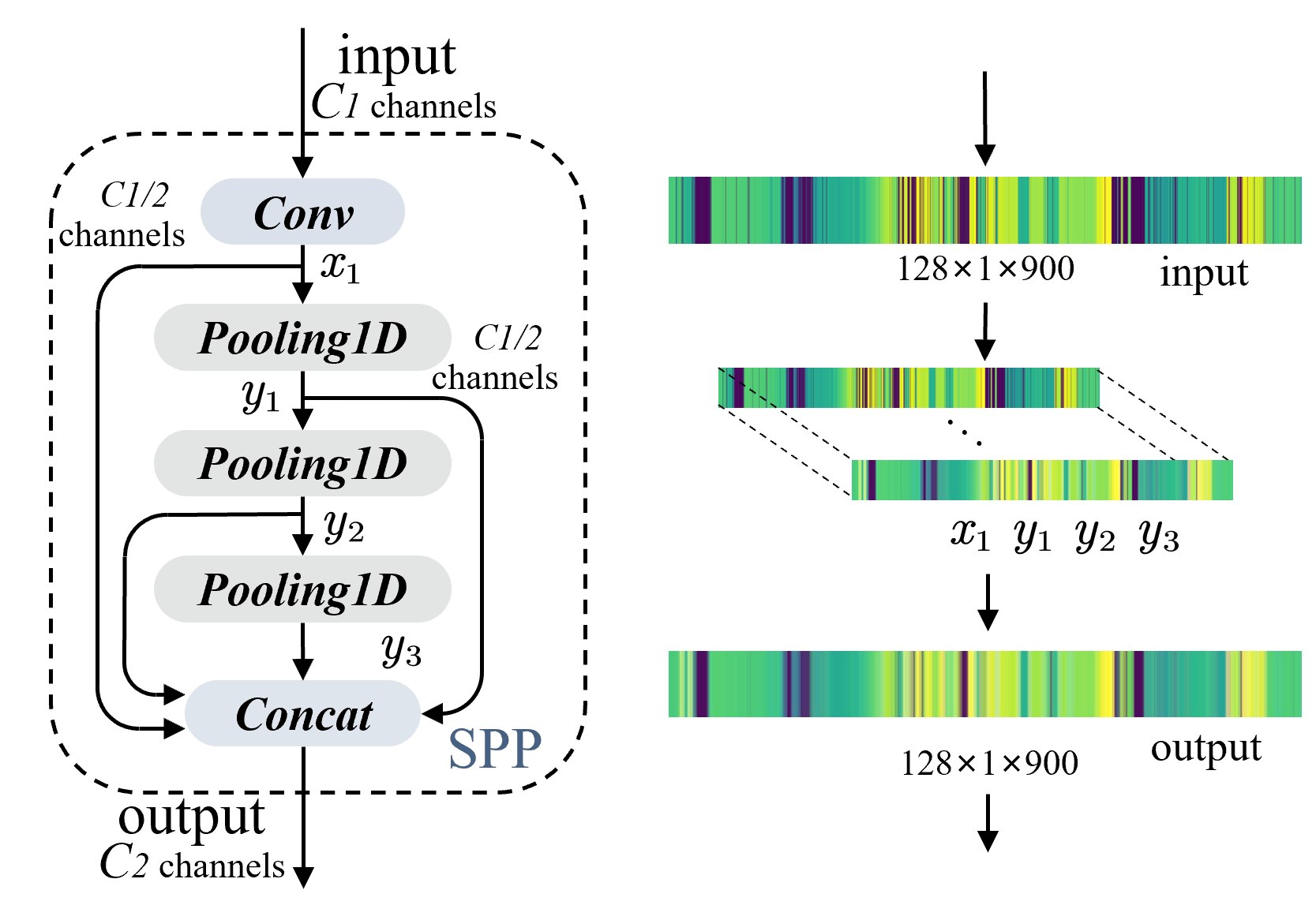}
\caption{The structure of the SPP block. Three consecutive 1D pooling operations are performed in the block, and the intermediate states are added together to obtain the output. A simple example is used to visualize the processing of the sequence, demonstrating that the SPP preserves yaw information while enriching the originally discontinuous spatial information.}
\label{fig3}
\end{figure}

We explored two distinct approaches for token sequence generation to ensure yaw equivariance. The first approach utilizes flattened 2D patches enhanced with positional encoding. The second adopts a purely convolutional framework. The distance image size is 64\(\times\) 900, which is closer to the representation of sequence data than normal images of equal sizes in the vertical and horizontal directions. Naturally, more feature information can be retained after converting to a sequence. However, methods for processing small patches and combining positional encoding have been proposed for normal images. When applied to distance images, an imbalance exists in the feature information in the horizontal and vertical directions. The overlap backbone uses convolutional filters along the vertical direction to compress the distance image into a feature sequence of size \(1 \times w \times c\) to address this issue.

This method generates a sequence that maximally preserves yaw information along the width dimension. Additionally, since the distance image has only 64 pixels in the vertical dimension, the backbone does not require large filters or many stacked convolutional modules. Overall, this method is more suitable for processing RVs.

In processing the range image, since the filters compress only the image along the vertical direction, the image is divided into \(H\) sequences of length \(W\) along the horizontal dimension for processing. However, the resulting feature sequences may exhibit incorrect spatial information due to object distortions and noise interference inherent in distance images. Therefore, we propose a simple architecture for the SPP module inspired by spatial pyramid pooling~\cite{he2015spatial}, as shown in Figure~\ref{fig3}.

SPP adopts two layers of 1-D pooling along the horizontal dimension without using multiscale pooling kernels. It performs three consecutive max-pooling operations on the input sequence and concatenates the intermediate states, followed by channel compression using filters. The pyramid pooling structure is simple but generally cannot be used for sequence processing because it aims to learn multiscale features in 2D images. However, as mentioned earlier, the sequences generated by vertical convolutional processing contain all the positional information in the horizontal direction. Therefore, using SPP can effectively improve the invariance of object positions and proportions in the sequence and reduce feature loss caused by noise interference.

Figure~\ref{fig3} shows that the input sequence undergoes two consecutive max-pooling operations, processing the sequence at different resolutions to capture richer spatial information. The processed sequence changes uniformly while maintaining the spatial hierarchical structure.

\subsection{Improved Triplet Loss with Hard Mining}\label{sec3.5}

According to \cite{r29,ma2023cvtnet}, using overlap to supervise the network is a suitable method. The degree of overlap is calculated by setting an overlap threshold to judge the similarity between two LiDAR scans. During training, the model selects positive and negative samples based on the overlap values and integrates them into a training batch. For each training tuple, we calculate the triplet loss using a query global descriptor \(g_q\), \(k_p\) positive descriptors \(g_p\), and \(k_n\) negative descriptors \(g_n\).

In traditional designs, the triplet loss is often calculated using the average distance between \(g_q\) and \(g_p\) and between \(g_q\) and \(g_n\). This aims to learn more subtle features from these positive and negative samples with small differences. The formula is as follows, where \(a\) represents the margin, \(\left\| . \right\| _{2}^{2}\) calculates the squared Euclidean distance, \(+\) indicates that the value is the loss when greater than 0, and the loss value is 0 when less than 0.

\begin{equation}
\begin{aligned}
&\mathcal{L} \left( g_{q},\left\{g_{p}^{}\right\},\left\{g_{n}^{}\right\} \right) =\\
&\sum_{k=1}^N\left[\left\|  g_{q}^{}  - g_{p}^{}  \right\| _{2}^{2}- \left\|  g_{q}^{}  - g_{n}^{} \right\|_{2}^{2}+\alpha \right] _+
\end{aligned}
\end{equation}

\begin{figure}[!t]
    \centering
    \includegraphics[width=0.85\linewidth]{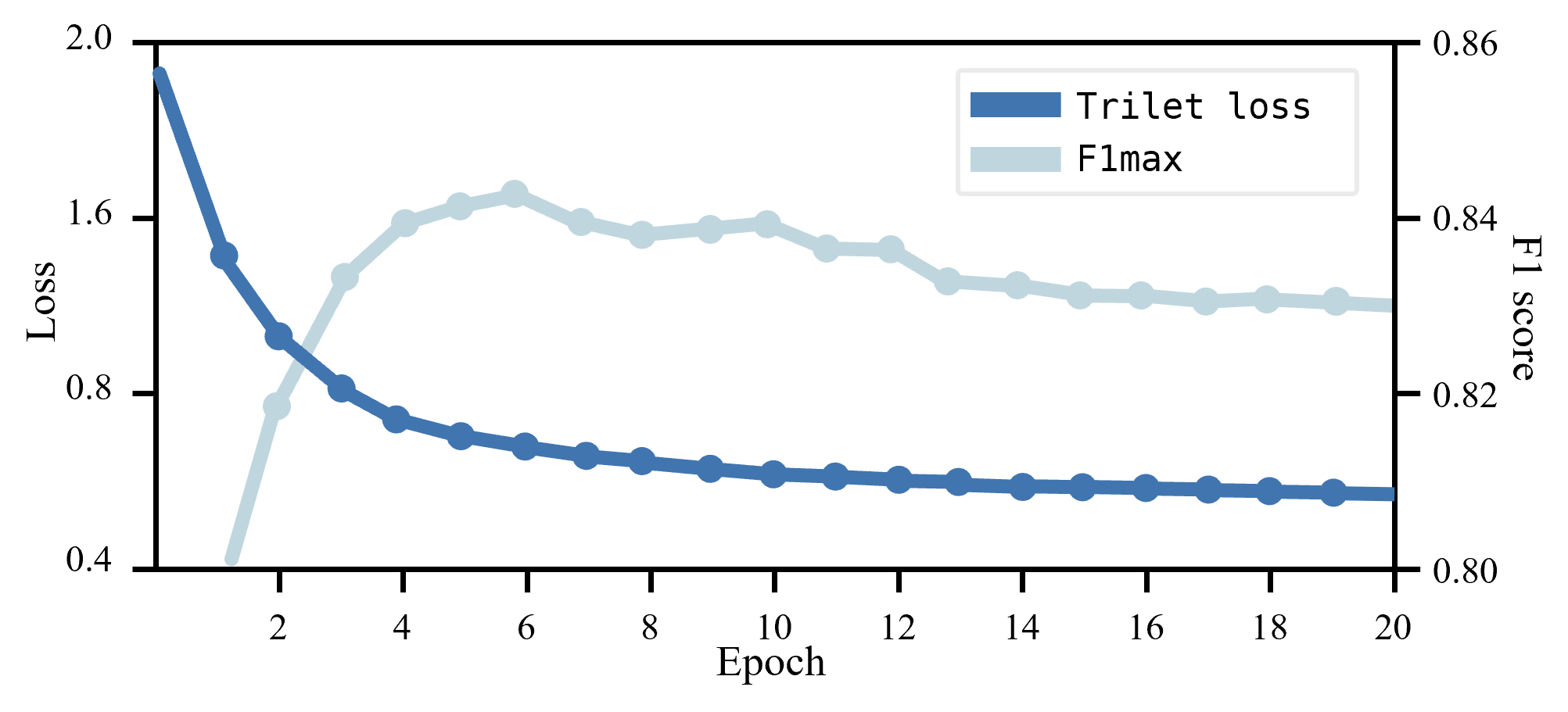}
\caption{Original loss and F1max during training. The model does not learn enough generalized information as the loss converges rapidly. The figure shows that the model overfit before the loss fully converged.}
\label{fig4}
\end{figure}

In our experiments, we found that the loss function had difficulty converging when using the traditional triplet loss. Additionally, the generalization ability of the model did not increase with the decreasing loss function, as shown in Figure~\ref{fig4}. We believe that the model's selected training data distribution is not uniform. The practice of randomly selecting samples from training data, although simple, results in easily differentiated samples. Many easily distinguishable samples are not conducive to the network learning better representations. To address this issue, we propose selecting positive and negative samples with the largest differences from the query sample for training, as shown in the following formula.

\begin{equation}
\begin{aligned}
&\mathcal{L} _T\left( g_q,\left\{ g_p \right\} ,\left\{ g_n \right\} \right) =\\
&\lambda d\left( g_q,g_p \right) +k_p\left( \alpha +\max_p \left( d\left( g_q,g_p \right) \right) \right) -k_n\min_n \left( d\left( g_q,g_n \right) \right)\\
\end{aligned}
\end{equation}

$\lambda$ represents a compression coefficient, and $d(\cdot)$ is used to compute the squared Euclidean distance. We use the triplet loss to minimize the distance between the query descriptor and the hardest positive global descriptor and to maximize the distance between the query and the hardest negative global descriptor while adding the distance between the query descriptor and the positive sample to ensure the absolute distance between the query and the positive sample. $\lambda$ controls the strength of this supervision term and is set to $1\times10^{-4}$ by default during training.

\section{Experiments}\label{sec4}

\begin{table}[!t]
\caption{Comparison of loop closure detection performance in KITTI and Ford Campus dataset}\label{tab1}
\renewcommand\arraystretch{1.5}
\begin{tabular}{p{0.6cm}lccccc}
\toprule
\textbf{Dataset} &\textbf{Approach}  &\textbf{AUC} &\textbf{F1max} &{\makecell[c]{\textbf{Recall}\\ \textbf{@1}}} & {\makecell[c]{\textbf{Recall}\\ \textbf{@1\%}}} \\
\midrule
 & Scan Context \cite{r4}   &0.836 &0.835 &0.820 &0.869   \\
    & Histogram \cite{r16}   &0.826 &0.825 &0.738 &0.871   \\
    & LiDAR Iris \cite{r18}  &0.843 &0.848 &0.835 &0.877 \\
    & PointNetVLAD \cite{Uy_Lee_2018}  &0.856 &0.846 &0.776 &0.845 \\
  {KITTI}  & OverlapNet \cite{r29}  &0.867 &0.865 &0.816 &0.908 \\
    & OverlapTransformer \cite{r30}  &0.907 &0.877 &0.906 &0.964 \\
    & NDT-Transformer-P \cite{r49}  &0.855 &0.853 &0.802 &0.869\\
    & MinkLoc3D \cite{r15}  &0.894& 0.869 &0.876 &0.920 \\
    & CVTNet \cite{ma2023cvtnet}  &0.911 &0.880 &- &-\\
    & \textbf{OverlapMamba(Ours)} &\textbf{0.934} & \textbf{0.890} & \textbf{0.898} & \textbf{0.959} \\
\midrule
    & Scan Context \cite{r4} 	&0.903	&0.842	&0.878	&0.958\\
    & Histogram \cite{r16}  &0.841	&0.800	&0.812	&0.897\\
    & LiDAR Iris \cite{r18} 	&0.907	&0.842	&0.849	&0.937\\
   {Ford } & PointNetVLAD \cite{Uy_Lee_2018} 	&0.872	&0.830	&0.862	&0.938\\
    {Campus}&OverlapNet \cite{r29}	&0.854	&0.843	&0.857	&0.932\\
      & OverlapTransformer \cite{r30} &0.923	&0.856	&0.914	&0.954\\
    & NDT-Transformer-P \cite{r49} 	&0.835	&0.850	&0.900	&0.927\\
    & MinkLoc3D \cite{r15} 	&0.871	&0.851	&0.878	&0.942\\
    &\textbf{Ovelap Mamba(Ours)}   & \textbf{0.929} &\textbf{0.871}  & \textbf{0.897} & \textbf{0.957} \\
\bottomrule
\end{tabular}
\end{table}

\subsection{Experimental Setup}\label{sec4.1}
We evaluate our method using three datasets: KITTI ~\cite{geiger2012we}, Ford Campus ~\cite{pandey2011ford}, and the publicly available NCLT~\cite{carlevaris2016university}. The KITTI dataset contains real image data collected in urban, rural, and highway scenes, with up to 15 cars and 30 pedestrians in each image, along with various degrees of occlusion and truncation. The Ford Campus Vision and LiDAR dataset is collected from an autonomous ground vehicle testing platform. The vehicle trajectory paths in these datasets contain several large and small-scale loops. The NCLT dataset was collected at the University of Michigan, where the dataset repeatedly explores the campus, both indoors and outdoors, on varying trajectories and at different times of the day across all four seasons. We use range images of size 1 \(\times\)  64 \(\times\)  900 for the KITTI and Ford datasets. For the NCLT dataset with 32-beam LiDAR data, we generate range images of size 1 \(\times\)  32 \(\times\)  900 containing all LiDAR points within 60 meters. To ensure a fair comparison with recent research, we set the maximum distance for the KITTI and Ford sequences to 50 m and similarly generated all point clouds in single-channel range images.

For the experiments with OverlapMamba, we use a single layer of the OverlapMamba block, where the embedding dimension is \(d_{model} = 256\). We sum the processed and unprocessed sequences one-to-one for the random yaw augmentation of the feature sequence. In the SPP module, we set the pooling kernel size to 5 and pad the sequences accordingly to maintain the length. We use the Adam optimizer~\cite{kingma2014adam}, with a learning rate initialized to \(5 \times 10^{-6}\), and train the OverlapMamba model for 20 epochs. We also attempt training with only LiDAR point cloud data, without using any other information, and generalize the results to different environments without fine-tuning.


\begin{figure}[!t]
    \centering
    \vspace{+.2cm}
    \includegraphics[width=0.9\linewidth]{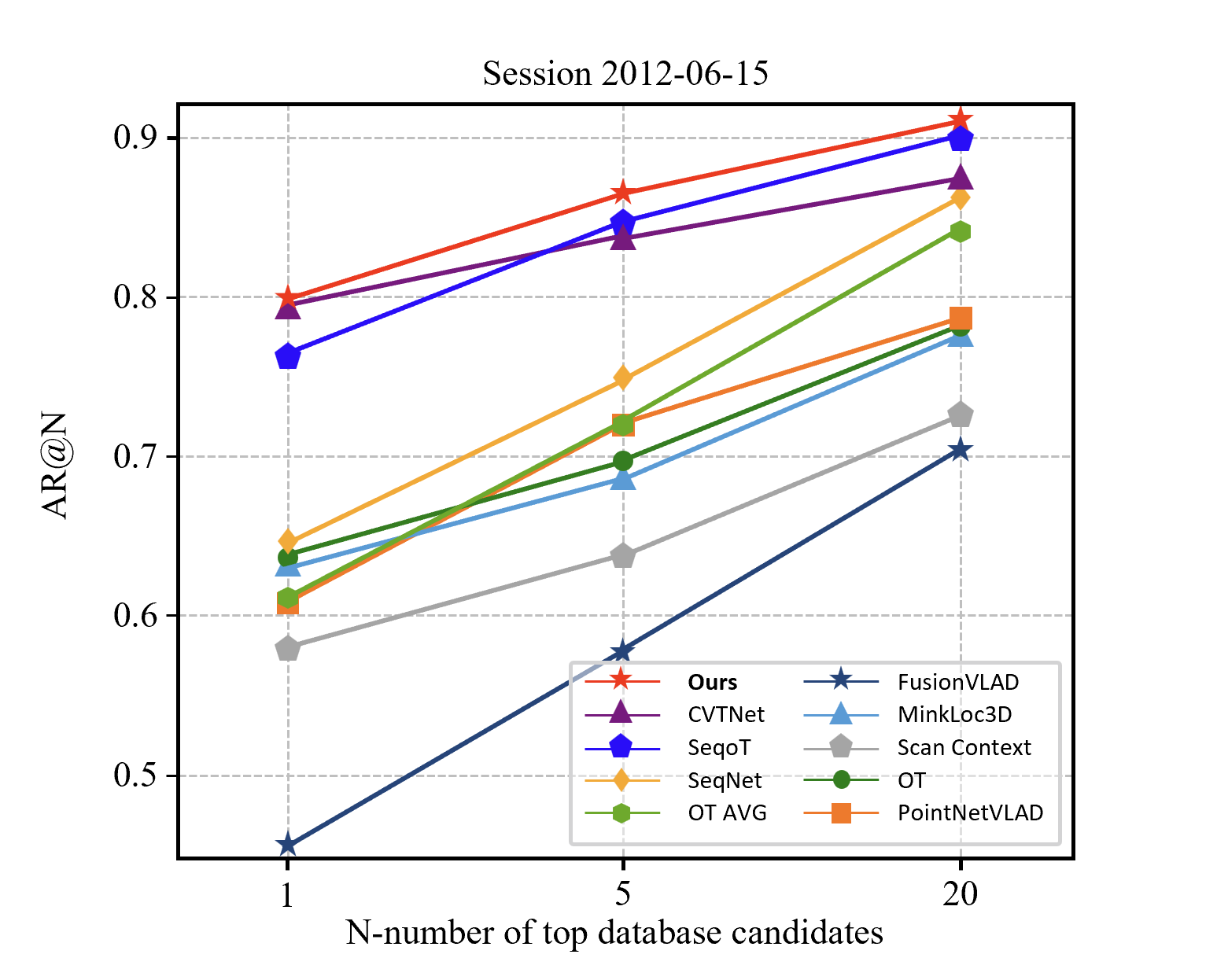}
\caption{Place recognition results using the NCLT dataset session 2012-06-15 as the query and 2012-01-08 as the database.}
\label{fig6}
\vspace{-.cm}
\end{figure}

\subsection{Evaluation for Loop Closure Detection}\label{sec4.2}

We trained and evaluated our method on the KITTI dataset and compared its performance with that of other baseline learning methods. We adopted the same experimental setup, training on sequences 03$\sim$10 and evaluating on sequences 00 and 02. We calculated the overlap value between two scans and scans with an overlap greater than the threshold of 0.3 were considered loop closures. Before training, we set the maximum number of positive and negative samples to 6 each.

The first experiment supports our claim that our method achieves state-of-the-art localization and loop closure detection in large-scale outdoor environments using LiDAR data and can generalize well to different environments. We evaluate the AUC, F1max score, recall@1, and recall@1\%, and the results are shown in Table~\ref{tab1}. OverlapMamba, trained only on depth range images, outperforms all methods based on point cloud features on KITTI. Compared to methods based on visual features, there is a 2.5\% and 1.3\% improvement in the F1 score over those of OverlapNet and OverlapTransformer, respectively. Compared to CVTNet, which uses RV and BEV combined inputs, OverlapMamba achieves a 2.3\% greater AUC and a 1\% greater F1max on the KITTI dataset. Our recall@1 and recall@1\% are among the highest and are comparable to those of OverlapTransformer, which incorporates transformer encoders. However, the overall data indicate that our model has better generalization capabilities. This is further evident in the comparison on the Ford dataset. We test on the Ford datase using the weights trained on KITTI sequences 03-10. The results show that our model achieves the highest F1max level of 0.871 on the Ford dataset, which is an improvement of nearly 2\% over the best existing method, OverlapTransformer. The results show that our proposal maintains consistent superiority on the test set. Our performance is superior to that of all the other methods.

\begin{figure}[!t]
    \centering
    \vspace{+.4cm}
    \includegraphics[width=0.9\linewidth]{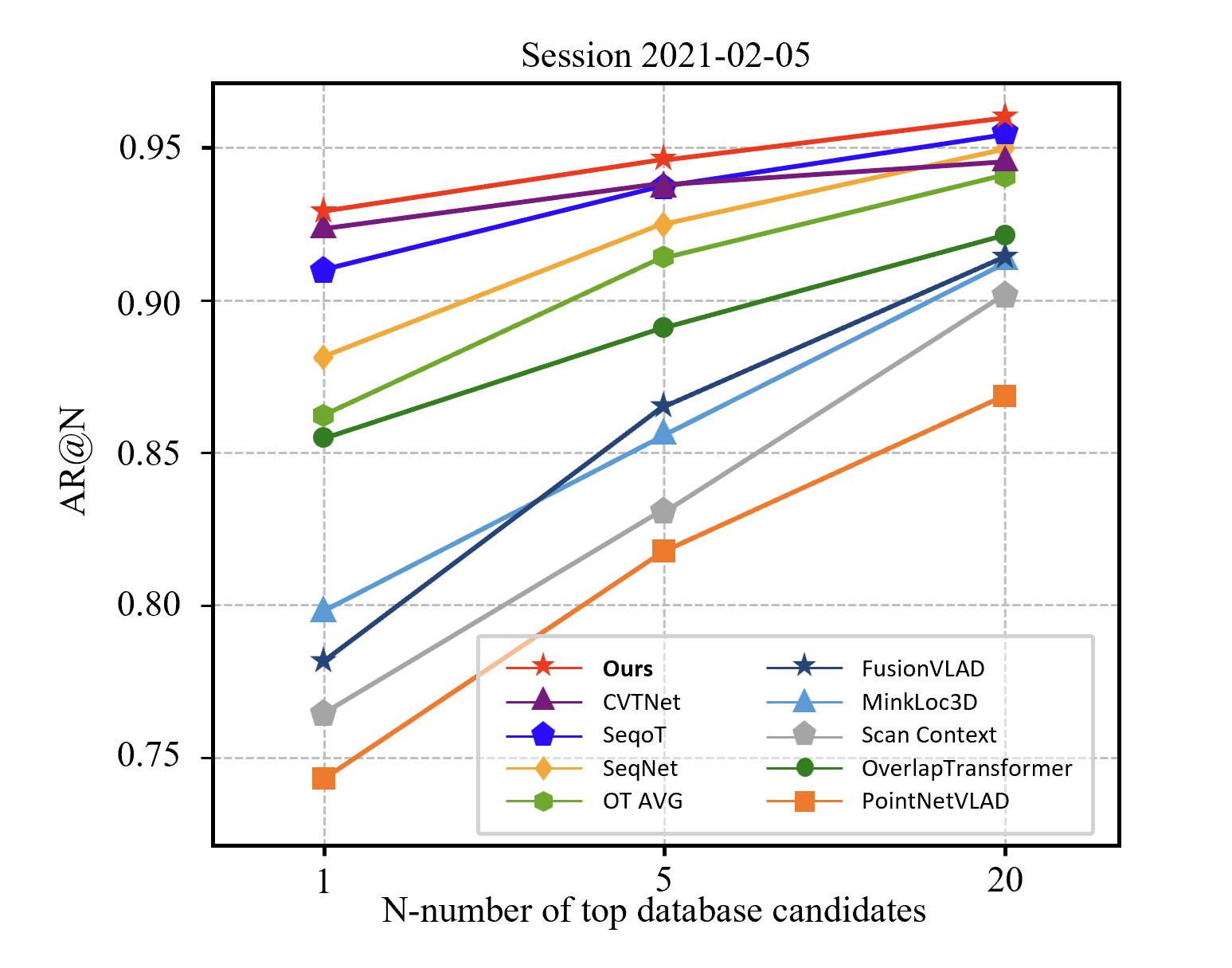}
\caption{Place recognition results using the NCLT dataset session 2012-02-05 as the query and 2012-01-08 as the database.}
\label{fig5}
\vspace{-.cm}
\end{figure}

\subsection{Evaluation for Place Recognition}
\label{sec4.3}
In this experiment, we validate the performance of our method on the NCLT dataset and compare it with other baseline methods. Similar to all baseline methods, we train our model using the database from 2012-01-08 and evaluate its performance on query sequences from 2012-02-05 and 2012-06-15. We use the same training settings as in the loop closure detection experiment. After training, we traverse the query sequences with a step size of 5. We iterate through all database sequences with a step size of 1 for each traversed query sequence, generating ground truth files for location recognition based on the Euclidean distance.

To accurately evaluate the model's performance, we use the average top 1 recall (AR@1), top 5 recall (AR@5), and top 20 recall (AR@20) as evaluation metrics for place recognition on the NCLT dataset. The evaluation results, as shown in Figure~\ref{fig6} and Figure~\ref{fig5}, indicate that our proposed method, OverlapMamba, significantly outperforms all baseline methods on the NCLTRemark dataset. Remarkably, we improve the AR@1 by 0.43\%$\sim$1.30\% and the AR@20 by 0.63\%$\sim$4.13\% compared to the last SoTA CVTNet \cite{ma2023cvtnet}, showcasing the significant potential of the Mamba architecture for long-term modeling tasks. OverlapMamba uses only RVs as input but outperforms all the baseline methods that employ diverse data representations, such as those that use only BEVs or combine RVs and BEVs as input, validating the superiority of our method in data processing and place recognition tasks.

\subsection{Ablation Study on Mamba Modules}\label{sec4.4}
We conducted ablation experiments on the proposed OverlapMamba and its different components; the results are shown in Table~\ref{tab2}. We used OverlapTransformer as the baseline. First, without any refinement modules, the method of using Mamba to process global descriptors was proven to be superior. Compared to OverlapTransformer, which uses a transformer as the main encoder structure, Mamba shows performance improvements that are similar to or surpass those of the transformer with only the linear complexity (setting $i$). Additionally, each proposed component improved the performance of the baseline to varying degrees (setting $ii$). Ablation experiments were conducted on the combinations of components $iii$ to demonstrate the indispensability of each component. The last row shows that OverlapMamba has the best performance for each component.

The experiments in Table~\ref{tab3} explored the effectiveness of the number of OverlapMamba modules in generating global descriptors. We modified the number of stacked modules in the OverlapMamba block and compared only 1$\sim $3 stacked modules. The experimental results on the KITTI dataset are shown in Table~\ref{tab3}, indicating that the best performance is achieved when using only a single Mamba module. This may be because when multiple Mamba blocks are stacked, more iterations and training samples are needed to achieve better performance. Therefore, using a single-layer OverlapMamba block can achieve an optimal balance between accuracy and runtime.

\begin{table}[t]
\caption{Ablation Experiments with Proposed Modules on KITTI Dataset.}\label{tab2}
\centering
\resizebox{\columnwidth}{!}{%
{\begin{tabular}{cccccc}
    \toprule
    Methods & \multicolumn{3}{c}{Component} & AUC & F1max\\
    \midrule
    & Shift & SPP & ImTrihard Loss &  \\
    Baseline  & \multicolumn{3}{c}{} & 0.891 & 0.842 \\
    OverlapMamba (\textit{i}) & - & - & - & 0.898 & 0.843  \\
    \midrule
    \multirow{3}{*}{OverlapMamba (\textit{ii})} & \checkmark & - & - & 0.901 & 0.848  \\
    & - & \checkmark & - & 0.882 & 0.845 \\
    & - & - & \checkmark & 0.881 & 0.850 \\
    \midrule
    \multirow{4}{*}{OverlapMamba (\textit{iii})} & \checkmark & \checkmark & - & 0.930 & 0.872 \\
    & \checkmark & - & \checkmark & 0.913 & 0.858  \\
    & - & \checkmark & \checkmark & 0.926 &0.857 \\ 
    & \checkmark & \checkmark & \checkmark & \textbf{0.934} & \textbf{0.890} \\
\bottomrule
\end{tabular}}}
\end{table}

\begin{table}[t]
\centering
\caption{Comparison of different numbers of OverlapMamba blocks on the KITTI dataset.}\label{tab3}
\begin{tabular}{cccc}
    \toprule  
     Number & Runtime(ms) & AUC  & F1max\\
    \midrule
       1 & 5.1  & 0.934 & 0.890   \\
       2 & 7.8  & 0.848 & 0.803   \\
       3 & 10.4  & 0.822 & 0.782   \\
\bottomrule
\end{tabular}
\label{comparation}
\end{table}

\subsection{Study on ImTrihard Loss}\label{sec4.5}
\begin{figure}[t]
    \centering
    \vspace{+.2cm}
    \includegraphics[width=0.9\linewidth]{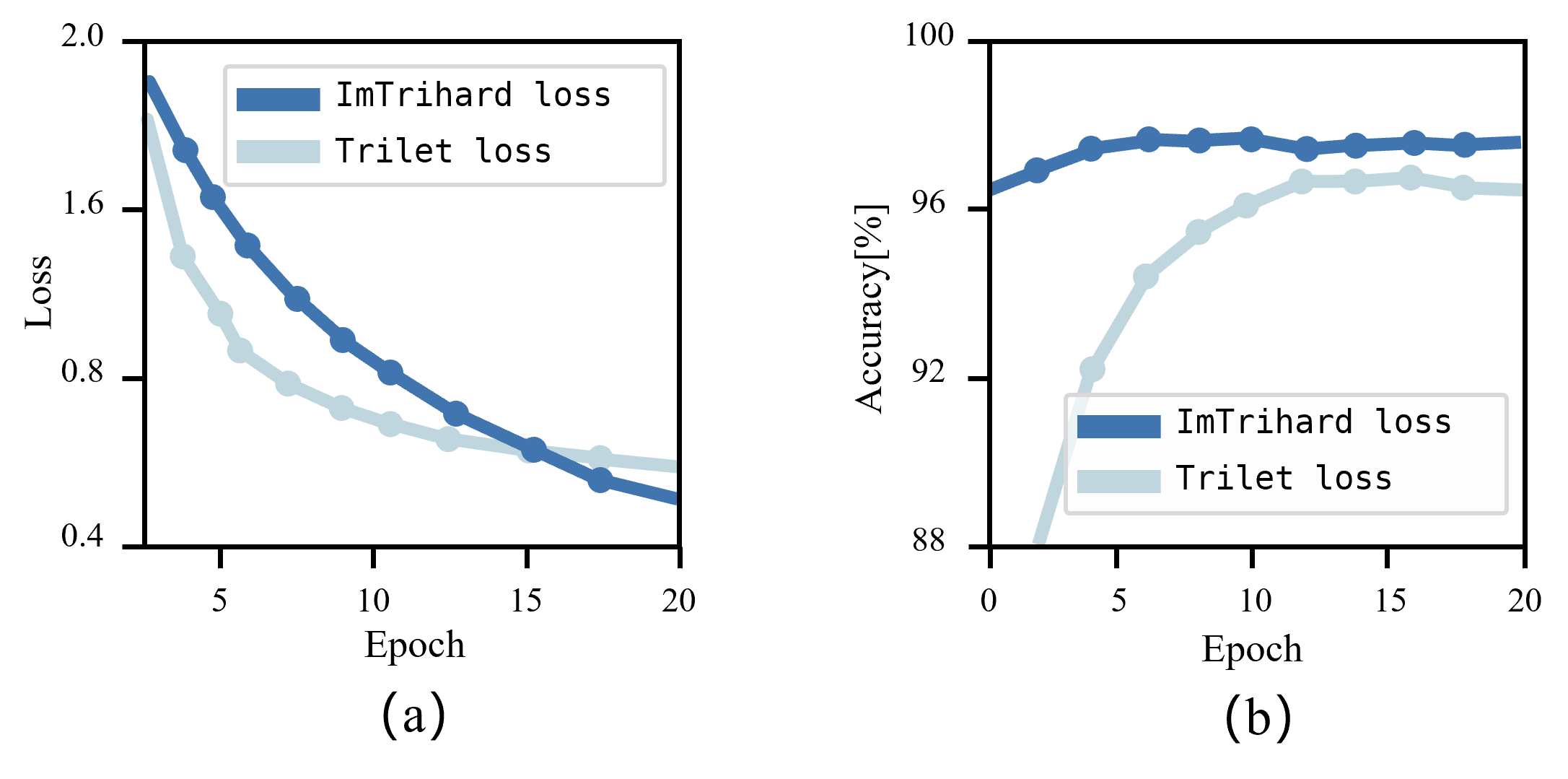}
\caption{Comparison of two loss functions. (a) shows the change in the loss value, and (b) shows the evaluation on the sequence 00 of the KITTI dataset.}
\label{fig7}
\vspace{-.cm}
\end{figure}

In Section~\ref{sec3.5}, we described the effect of our proposed ImTrihard loss function on the model's generalization ability and training convergence speed. We conducted experiments on the KITTI dataset (Table~\ref{tab4}) to further validate the reduction in training time cost achieved by our ImTrihard loss. Notably, in just the first training epoch, the F1max score of the OverlapMamba trained with ImTrihard loss reached 0.872, surpassing the F1 scores of many existing methods after full training. Moreover, because ImTrihard loss selects the hardest positive and negative samples, its loss value may initially be greater than that of traditional triplet loss. In the experiments, we also observed that triplet loss exhibited signs of overfitting as the number of training epochs increased. In the 20th training epoch, F1max was approximately 1.2\% lower than F1max in the 10th epoch. However, no performance decline was observed in the experiments using ImTrihard loss. The change in loss values is depicted more intuitively in Figure~\ref{fig7}. The evaluation on sequence 00 of the KITTI dataset (b) further demonstrates its convergence capability. In the first epoch alone, its accuracy reached 96.43\%. These results further confirm that the ImTrihard loss can help the model converge quickly and has excellent generalizability. Its relatively simple structure also facilitates its easy application to different scenarios.

\begin{table}[t]
\centering
\caption{Comparison of convergence speed in the training process of OverlapMamba on two loss functions.}\label{tab4}
\begin{tabular}{cp{1.5cm}p{1.5cm}p{1.5cm}p{1.5cm}}
    \toprule  
    &\multicolumn{2}{@{}l}{Trilet loss (original)}&\multicolumn{2}{@{}l}{ImTrihard loss} \\
\cmidrule{2-3}\cmidrule{4-5}
     Epoch & Loss   & F1max & Loss   & F1max\\
    \midrule
       1 & 1.231  & 0.776 & 1.925 & 0.872  \\
       5 & 0.803  & 0.826 & 1.431 & 0.880  \\
      10 & 0.667  & \textbf{0.844} & 1.003 & 0.888  \\
      20 & 0.571  & 0.832 & 0.557 & \textbf{0.890} \\
\bottomrule
\end{tabular}
\label{comparation}
\end{table}

\subsection{Runtime}\label{sec4.6}
In this experiment, we use the NCLT sequence 2012-01-08 as the database and the sequence 2012-02-05 as the query to calculate the runtime. Since the Mamba architecture has the advantages of fast inference, OverlapMamba outperforms all the baseline methods because of its efficient SSM scan, which is up to 5 times faster than that of a transformer of similar size. As shown in Table \ref{tab5}, our method is the fastest among the SoTA methods. For descriptor generation, with 0.49 ms for each scan, even faster than the pure geometric histogram-based method. For the time cost of searching, our method significantly reduces the runtime, requiring only 0.35 ms on average. OverlapMamba performs a straightforward calculation of Euclidean distances between descriptors during the search process without any preprocessing or calculating the similarity of descriptors through functions. Specifically, compared with the same descriptor size of 1  \(\times\)  256 generated by the baseline method, our approach demonstrates the descriptive power of the generated descriptors and its unparalleled real-time performance.

\begin{table}[t]
\centering
\caption{Comparison of runtime with state-of-the-art methods.}\label{tab5}
\begin{tabular}{clcc}
    \toprule  
     \multicolumn{2}{c}{Approach} & \makecell{Descriptor \\ Extraction [ms]}  & Searching [ms] \\
    \midrule
       \multirow{3}{*}{\makecell{Hand \\ crafted}} & Histogram \cite{r16}                                & 1.07 & 0.46   \\
       & Scan Context \cite{r4}     & 57.95 & 492.63   \\
       & LiDAR Iris \cite{r18}      & 7.13 & 9315.16   \\
    \midrule
    \multirow{7}{*}{\makecell{Learning \\ based}}& PointNetVLAD \cite{Uy_Lee_2018}                  & 13.87 & 1.43   \\
       & OverlapNet \cite{r29}          & 4.85 & 3233.30   \\
       & NDT-Transformer-P \cite{r49}   & 15.73 & 0.49   \\
       & MinkLoc3D \cite{r15}           & 15.94 & 8.10   \\
       & OverlapTransformer \cite{r30}  & 1.37 & 0.44   \\
       & Ours                           & \textbf{0.49} & \textbf{0.35}   \\
\bottomrule
\end{tabular}
\end{table}

\section{Conclusion}
In this paper, we propose a novel LiDAR-based localization network which leverages the Mamba model, a random reconstruction method to process RVs, and a straightforward SPP architecture. Extensive experiment results prove that our proposed OverlapMamba can outperform other SOAT algorithms on three public datasets in time accurancy, complexity and speed even with simple information inputs, demonstrating its generalization ability in LPR task and practical value in real-world autonomous driving scenarios.

\bibliography{bibfile} 
\bibliographystyle{IEEEtran}

\begin{IEEEbiography} [
{\includegraphics[width=1in,height=1.25in,clip,keepaspectratio]{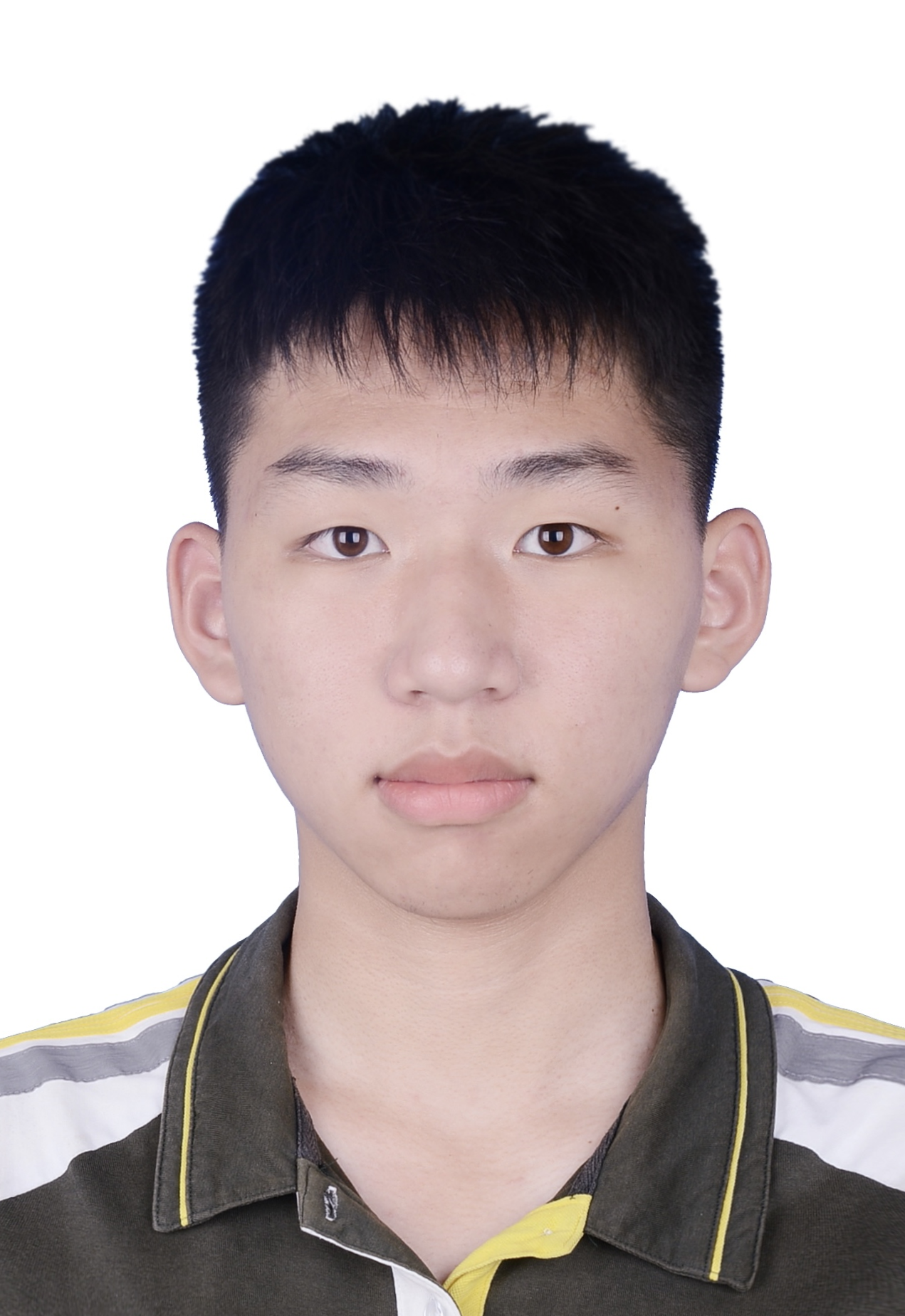}}]
{Qiuchi Xiang} was born in 2004. He is a student member of the Institute of Electrical and Electronics Engineers. He is currently pursuing a B.S. degree in data science and big data technology with Xingzhi College, South China Normal University Shanwei, China. His main research interests include computer vision and computer vision, SLAM, and deep learning.
\end{IEEEbiography}

\vspace{-0.4cm}

\begin{IEEEbiography} [
{\includegraphics[width=1in,height=1.25in,clip,keepaspectratio]{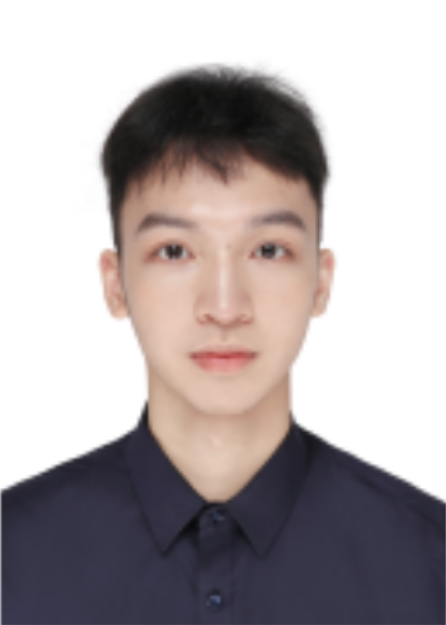}}]
{Jintao Cheng} received his bachelor's degree from the School of Physics and Telecommunications Engineering, South China Normal University, in 2021. His research includes computer vision, SLAM, and deep learning.
\end{IEEEbiography}

\begin{IEEEbiography} [
{\includegraphics[width=1in,height=1.25in,clip,keepaspectratio]{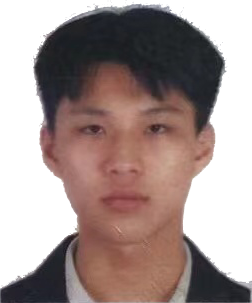}}]
{Jiehao Luo} is a student member of the Chinese Association for Artificial Intelligence. He is currently pursuing a B.S. degree in the Internet of Things with Xingzhi College, South China Normal University Shanwei, China. His main research interests include computer vision, IoTs and deep learning.
\end{IEEEbiography}
\begin{IEEEbiography}[
{\includegraphics[width=1in,height=1.25in,clip]{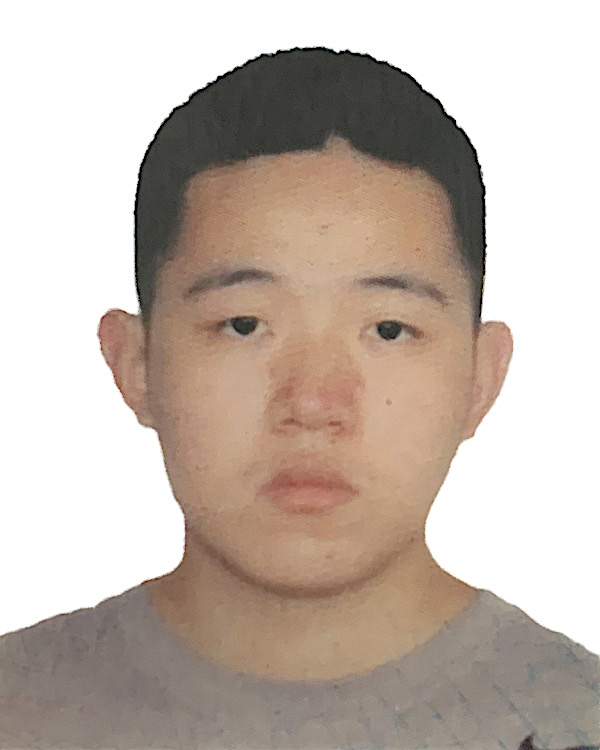}}]{Jin Wu}
(Member, IEEE) received a B.S. degree from the University of Electronic Science and Technology of China, Chengdu, China. From 2013 to 2014, he was a visiting student with Groep T, Katholieke Universiteit Leuven (KU Leuven). He is currently pursuing a Ph.D. degree in the Department of Electronic and Computer Engineering, Hong Kong University of Science and Technology (HKUST), Hong Kong, supervised by Prof. Wei Zhang. He has coauthored over 140 technical papers in representative journals and conference proceedings. He was awarded the outstanding reviewer of \textsc{IEEE Transactions on Instrumentation and Measurement} in 2021. He is now a review editor of \emph{Frontiers in Aerospace Engineering} and an invited guest editor for several JCR-indexed journals. He is also an IEEE Consumer Technology Society (CTSoc) member and a committee member and publication liaison. He was a committee member for the IEEE CoDIT in 2019, a special section chair for the IEEE ICGNC in 2021, a special session chair for the 2023 IEEE ITSC, a track chair for the 2024 IEEE ICCE, 2024 IEEE ICCE-TW and a chair for the 2024 IEEE GEM conferences. From 2012 to 2018, he was in the micro air vehicle industry and has started two companies. From 2019 to 2020, he was with Tencent Robotics X, Shenzhen, China. He was selected as the World's Top 2$\%$ Scientist by Stanford University and Elsevier in 2020, 2021 and 2022.
\vspace{10pt} 
\end{IEEEbiography}

\begin{IEEEbiography}[{\includegraphics[width=1in,clip]{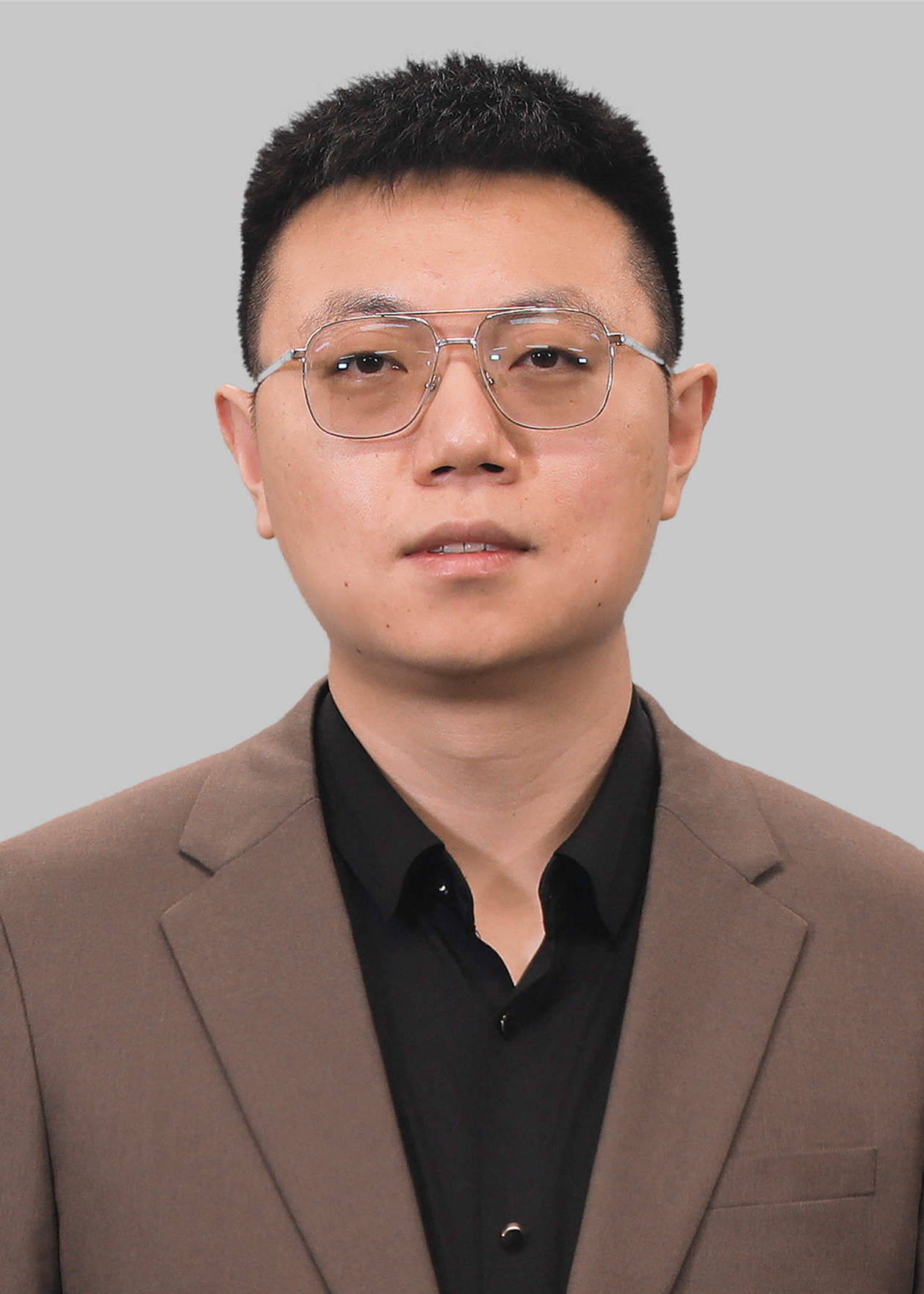}}]{Rui Fan}(Senior Member, IEEE) received a B.Eng. degree in automation from the Harbin Institute of Technology in 2015 and a Ph.D. degree (supervisors: Prof. John G. Rarity and Prof. Naim Dahnoun) in Electrical and Electronic Engineering from the University of Bristol in 2018. He worked as a research associate (supervisor: Prof. Ming Liu) at the Hong Kong University of Science and Technology from 2018 to 2020 and a postdoctoral scholar-employee (supervisors: Prof. Linda M. Zangwill and Prof. David J. Kriegman) at the University of California San Diego between 2020 and 2021. He began his faculty career as a full research professor with the College of Electronics \& Information Engineering at Tongji University in 2021 and was then promoted to a full professor in the same college and at the Shanghai Research Institute for Intelligent Autonomous Systems in 2022.

Prof. Fan served as an associate editor for ICRA'23 and IROS'23/24, an Area Chair for ICIP'24, and a senior program committee member for AAAI'23/24. He is the general chair of the AVVision community and has organized several impactful workshops and special sessions in conjunction with WACV'21, ICIP'21/22/23, ICCV'21, and ECCV'22. He was honored by being included in the Stanford University List of Top 2\% Scientists Worldwide in both 2022 and 2023, recognized on the Forbes China List of 100 Outstanding Overseas Returnees in 2023, and acknowledged as one of Xiaomi Young Talents in 2023. His research interests include computer vision, deep learning, and robotics.
\end{IEEEbiography}

\begin{IEEEbiography}[
{\includegraphics[width=1in,height=1.25in,clip,keepaspectratio]{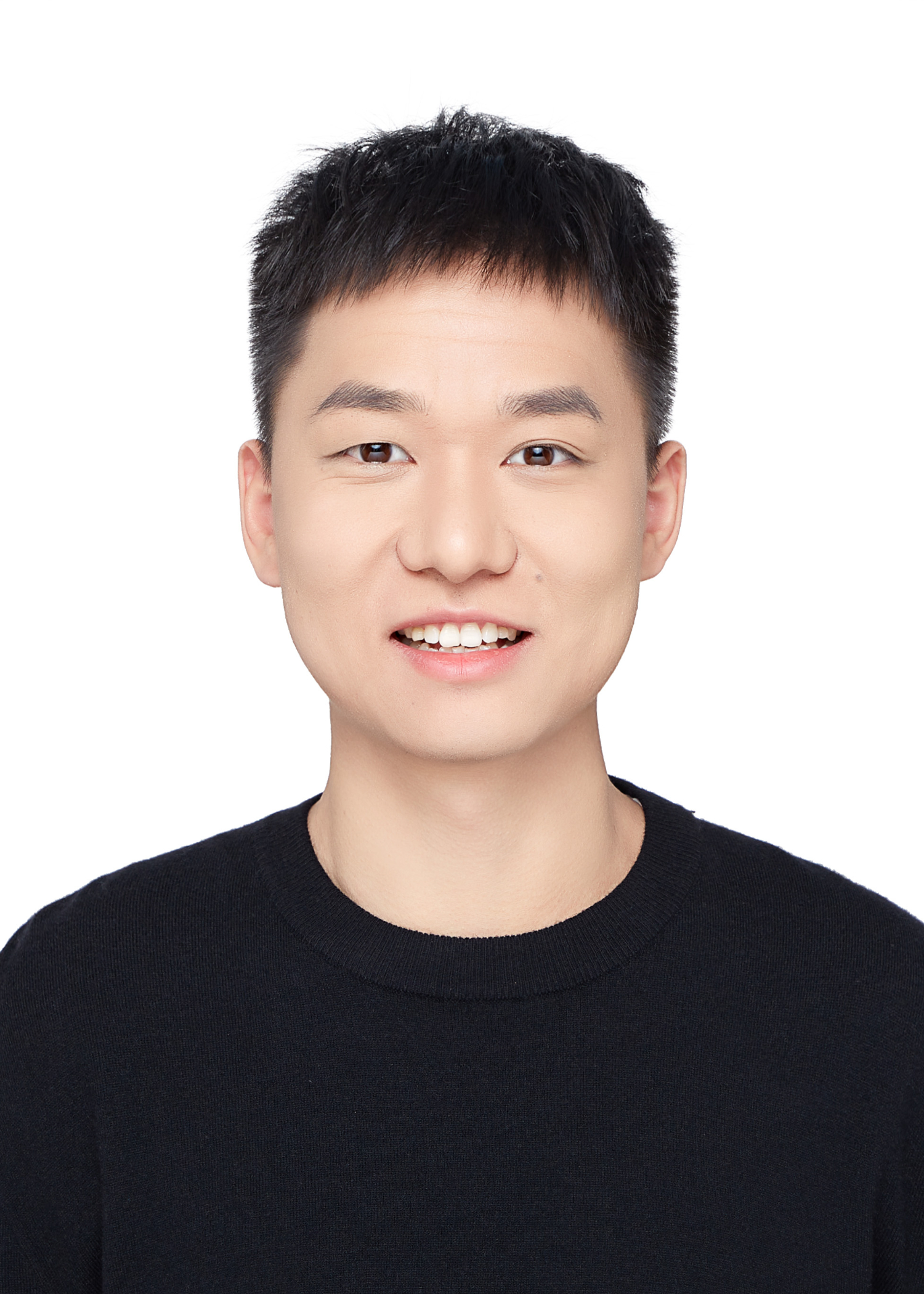}}]{Xieyuanli Chen}
Xieyuanli Chen is now an associate professor at the National University of Defense Technology, China. He received his Ph.D. degree in robotics at the Photogrammetry and Robotics Laboratory, University of Bonn. He received his master's degree in robotics in 2017 at the National University of Defense Technology, China, and his bachelor's degree in electrical engineering and automation in 2015 at Hunan University, China. He currently serves as an associate editor for IEEE RA-L, ICRA and IROS.
\end{IEEEbiography}

\begin{IEEEbiography}[
{\includegraphics[width=1in,height=1.25in,clip,keepaspectratio]{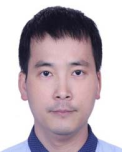}}]{Xiaoyu Tang}
(Member, IEEE) received a B.S. degree from South China Normal University in 2003 and an M.S. degree from Sun Yat-sen University in 2011. He is currently pursuing a Ph.D. degree at South China Normal University. He is working with the School of Physics, South China Normal University, where he is engaged in information system development. His research interests include machine vision, intelligent control, and the Internet of Things. He is a member of the IEEE ICICSP Technical Committee.
\end{IEEEbiography}

\vfill

\end{CJK}
\end{document}